# Digit Image Recognition Using an Ensemble of One-Versus-All Deep Network Classifiers

Abdul Mueed Hafiz[1][0000-0002-2266-3708] and Mahmoud Hassaballah[2][0000-0001-5655-8511]

[1]Department of ECE, Institute of Technology, University of Kashmir,
Srinagar, J&K, 190006, India
[2]Department of Computer Science, Faculty of Computers and Information,
South Valley University, Qena, 83523 Egypt
`mueedhafiz@uok.edu.in`

**Abstract.** In multiclass deep network classifiers, the burden of classifying samples of different classes is put on a single classifier. As the result the optimum classification accuracy is not obtained. Also training times are large due to running the CNN training on single CPU/GPU. However it is known that using ensembles of classifiers increases the performance. Also, the training times can be reduced by running each member of the ensemble on a separate processor. Ensemble learning has been used in the past for traditional methods to a varying extent and is a hot topic. With the advent of deep learning, ensemble learning has been applied to the former as well. However, an area which is unexplored and has potential is One-Versus-All (OVA) deep ensemble learning. In this paper we explore it and show that by using OVA ensembles of deep networks, improvements in performance of deep networks can be obtained. As shown in this paper, the classification capability of deep networks can be further increased by using an ensemble of binary classification (OVA) deep networks. We implement a novel technique for the case of digit image recognition and test and evaluate it on the same. In the proposed approach, a single OVA deep network classifier is dedicated to each category. Subsequently, OVA deep network ensembles have been investigated. Every network in an ensemble has been trained by an OVA training technique using the Stochastic Gradient Descent with Momentum Algorithm (SGDMA). For classification of a test sample, the sample is presented to each network in the ensemble. After prediction score voting, the network with the largest score is assumed to have classified the sample. The experimentation has been done on the MNIST digit dataset, the USPS+ digit dataset, and MATLAB digit image dataset. Our proposed technique outperforms the baseline on digit image recognition for all datasets.

**Keywords:** OVA Classification, Deep Ensemble Learning, MNIST; CNN.

## 1      Introduction

Despite significant success in the area of knowledge discovery, conventional machine learning approaches can fail to perform well when they deal with complex data if it is imbalanced, high-dimensional, noisy, etc. This is due to the difficulty of these techniques in capturing various features and underlying data structure [1]. Hence, an important research topic has evolved in data mining for effectively building an efficient knowledge discovery and mining model. Ensemble learning, being one research hotspot, aims at integration of fusion, modeling, and mining of data to form a unified model. Specifically, it extracts feature sets along with various transformations. Based on the features learnt, various algorithms produce weak predictions. Then, ensemble learning fuses the information obtained from the same for knowledge discovery and for better prediction with the help of adaptive voting [1].

Dong et. al. [1] presented an introduction to ensemble learning approaches. The conducted research in [2] used weak multiclass ensembles which combine outputs of different layer-based transfer conditions in deep networks. Experiments have shown that this reduces the effects of adverse feature transference of features in image recognition tasks [1]. The research done in [3] has also used weak multiclass ensembles to reduce the cross-domain error in domain adaptation for the task of sentiment analysis. The work in [4] has designed an ensemble which uses AdaBoost [5] for adjustment of weights of source data and target data. It achieved good performance on UCI datasets for insufficient data. Subsequently more work has been done in [6-14] which usually use ensembles of weak multiclass classifiers which do not give satisfactory results as will be discussed. One important type of ensemble which has potential and is unexplored is One-Versus-All (OVA) ensemble.

We propose to use a deep-learning [15] based OVA ensemble approach for digit image recognition. By doing this, an effort has been made to increase the classification accuracy of deep neural networks on the classical digit image recognition task by using them in an ensemble of binary-classification deep networks for the purpose of multi-class classification. Handwritten digit recognition is an applied and an interesting research area. Many approaches have been proposed for this, such as Convolutional neural network based classification techniques [16-18], reinforcement learning [19], artificial neural networks [20,21], support vector machines [22,23], etc. In spite of the fact that many these techniques have achieved decent classification accuracy on larger, more complex, and realistic images - issues remain due to issues including non-standard writing of circle and hook patterns, e.g. in 4, 5, 7 and 9, and also image perception issues such as skew, slant, blur, small size, etc. After applying the proposed approach to digit image recognition as demonstrated by the experiments, higher classification accuracy has been achieved as compared to that of conventional deep network classifiers and other state-of-art ensemble techniques on all the datasets used. Three digit image datasets namely the *MNIST digit image dataset* [24], the *USPS+ digit image dataset* [25] and the *MATLAB digit image dataset* were used. In one of the experiments which used a subset of the *MNIST digit image dataset*, the accuracy of Binary Classification Convolutional Neural Network (BCCNN) Ensemble was found to be 98.03% which was higher than that found after using a conventional Multi-class Convolutional Neural Network (MCNN) viz. 97.90%. After structural modifications to the ensemble, and subsequently testing it as well as a conventional deep network on a another subset of MNIST, it was found that the proposed technique gave an accuracy of 98.50% which was higher than that of the conventional deep network viz. 98.4875%.

The rest of the paper is structured as follows. Related works are discussed in Section 2. Section 3 discusses the proposed approach. In Section 4, experimentation details are discussed. We conclude in Section 5.

## 2. Related Works

In light of the previous works discussed in Section 1, we move to some notable works which we discuss below.

The authors in [9] state that neural networks lag behind other state-of-art algorithms for Time Series Classification (TSC). The latter approach is composed of an ensemble of 37 non deep-learning classifiers. The authors of this work attribute the lag to the absence of deep network ensembles in TSC. In this work, they illustrate that an ensemble of 60 deep networks significantly improves the state-of-the-art performance using deep networks for TSC. They have used the UCR/UEA archive [26] being the largest public database for time series analysis. Our work is about image recognition of digits and subsequently does not go in the direction of TSC.

The authors in [10] state that most deep networks suffer from difficulty of interpretation, and from overfitting. Although regularization has been investigated to avoid overfitting, but not much underlying theoretical analysis is present. In this work, it is argued that for extraction of neural network features for decision making, consideration of cluster paths in neural networks is important. The author of the work accordingly presents an ensemble of neural networks which gives good test accuracy. The technique gives state-of-the-art results for ResNets [27-29] on CIFAR-10 [30]. It also improves the performance of various models applied. However, CIFAR-10 images are too small for the task of general image recognition (which is the topic of the work) as the latter involves use of much larger images. Also using several weak classifiers may lead to decision overlap due to high correlation between the subsets of the database used for training the weak classifiers.

The authors in [6] state that sharing medical image datasets amongst different institutions is legally limited. Hence medical research requiring large datasets suffers a lot due to its requirement of immensely large imaging datasets. It introduces constrained Generative Adversarial Network ensembles (cGANe) for addressing this problem. These ensembles alter the imaging data while preserving information, enabling reproduction of research elsewhere with the shared data. However, the authors of this work state that their technique needs more validation for different medical image data types.

The authors in [31] state that as has been recently observed that Neural Machine Translation (NMT) models with deep networks can be more effective. However, these are difficult to train. The work presents a MultiScale Collaborative (MSC) framework for easing NMT model training. The models used are deeper than the earlier ones. They demonstrate that their MSC networks optimize easily and give quality improvements by considerably increasing the depth of the network. However, using very deep networks has the problems of training difficulty and parameter overloading.

As can be observed from above discussion, though deep ensemble learning is used in state-of-art, it suffers from problems associated with the techniques used. These in-clude but are not limited to narrow applicability, lesser investigation into the science of ensemble learning (leading to underutilization of the concept), alternative schemes, and implementations fraught with their own issues. The usual toll these issues take are low classification accuracy and long training time. In this paper we investigate the former, and show that by using deep ensemble learning in a new light, the state-of-art performance in ensemble learning can be advanced. We leave the investigation of speedup obtained by using deep OVA ensemble learning for future work.

## 3. Proposed Approach

A simple differentiation between the conventional deep multiclass classification and deep ensemble learning is shown in Figure 1. As can be seen from the figure, the ensemble technique uses a group of CNNs for voting for final decision. The OVA approach was used in multiclass classification using softmax aggregation of Binary SVM Classifiers [11] and good results have been obtained. Here, we extend the concept of multiclass OVA classification using prediction score voting in binary classification deep neural networks. To the best of our knowledge this is the first work to use an ensemble of OVA deep network classifiers which is unlike the state-of-art deep network ensembles found in literature.[1] The proposed approach uses deep neural networks as binary-classifiers in an OVA ensemble which increases their classification accuracy.

In the proposed technique, OVA approach is used for training as well as for classification. Each Binary Classification Convolutional Neural Network (BCCNN) in the ensemble has the same number of and type of layers as a conventional Multiclass Convolutional Neural Network (MCNN), but only up to and excluding the fully connected layer. In each BCCNN used, the fully connected layer consists of two neurons, followed by a softmax layer, which in turn is followed by a binary (sigmoid) classification layer. As per convention in a 10-digit image classifier MCNN, the fully connected layer consists of 10 neurons, followed by a softmax layer, which in turn is followed by a classification layer with ten neurons. The architectures of both the binary as well as the multiclass deep networks are shown in Figure 2.

The experimentation has been based on the task of digit-image recognition. $BCCNN_i$ (*i=1, 2, 3, ... 10)* is trained with the conventional deep network training algorithm of Stochastic Gradient Descent with Momentum. A 28-by-28 pixel array is used for each digit image. One-versus-all training is done end-to-end.

Once all BCCNNs have been trained, the ensemble is used for classification. A test sample to be classified is presented to all the 10 BCCNNs in a trained ensemble. Next, the prediction score of each BCCNN in the ensemble is monitored. The BCCNN with maximum prediction score is assumed to have classified the sample. It should be noted that the MCNN is trained and used as per convention.

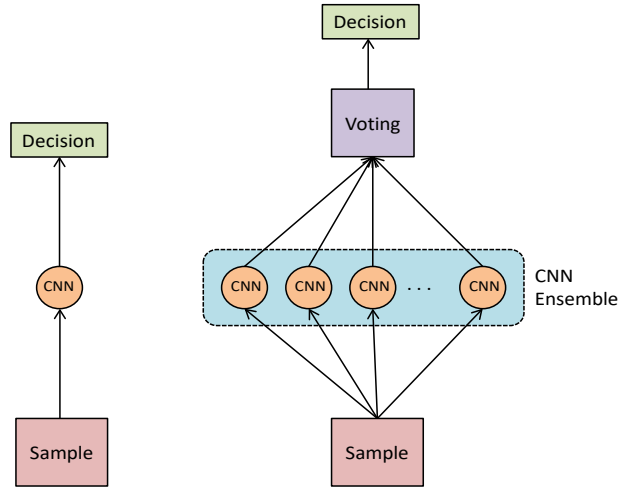

Figure 1. A comparison of different classification approaches.
(Left: Single multiclass CNN classification; Right: Deep ensemble classification)

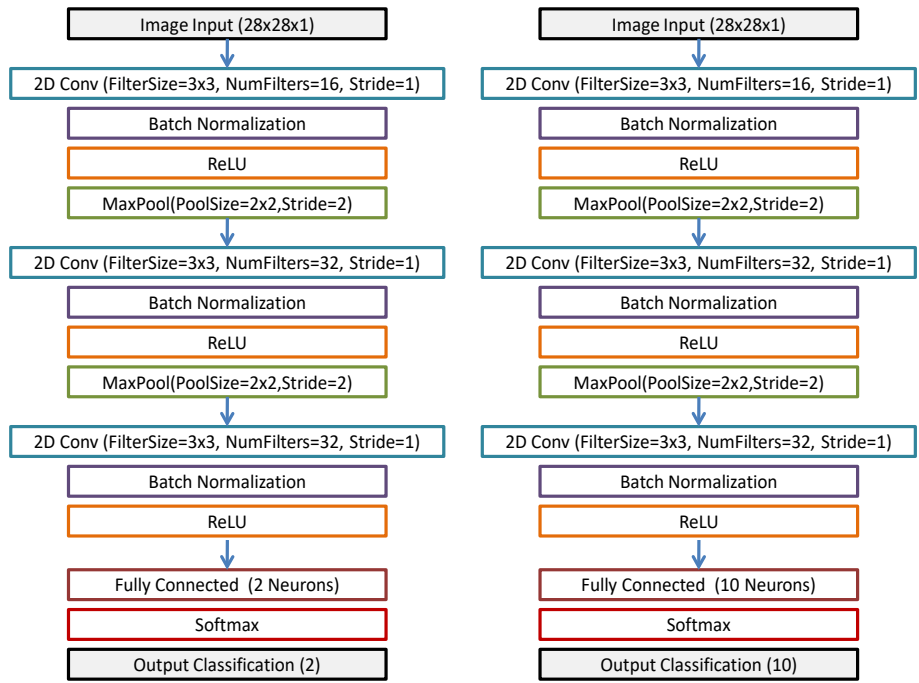

Figure 2. Deep network architectures used for experimentation. Left: A single CNN used in ensemble learning technique (BCCNN); Right: The multiclass CNN used (MCNN);

Figure 3 shows the overview of the proposed approach.

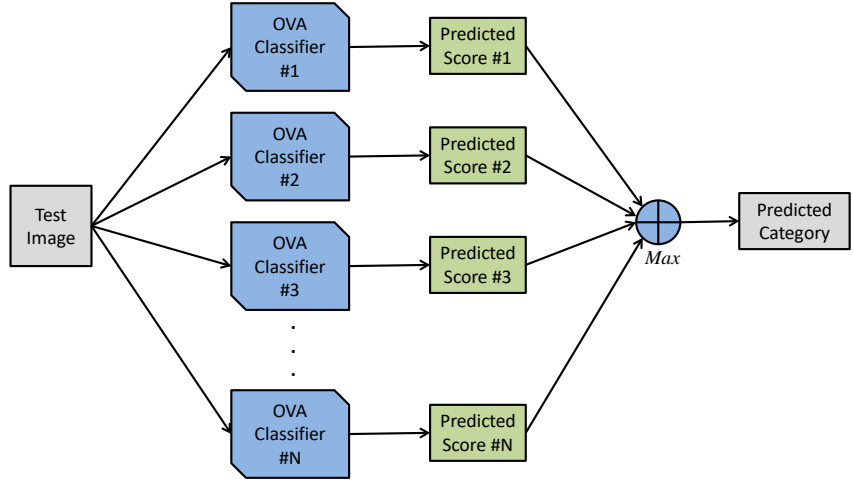

Figure 3. Overview of the proposed approach.

## 4. Experimentation and Results

Experimentation was done on an Intel CORE i3 processor system with 6GB of RAM running Windows7. Both the BCCNN ensemble, as well as the MCNN, were trained using the Stochastic Gradient Descent with Momentum Algorithm with *Initial Learn Rate* = 0.01, $L_2$ *Regularization Factor* = 0.0001, *Momentum* = 0.9, *Validation Frequency* = 50, and *Validation Patience* = 5. The digit-image datasets used were the *MATLAB Digit Image Dataset* (having 10000 instances), the *USPS+ Digit Image Dataset* (having 11000 instances) and the *MNIST Digit Image Dataset* (having 70000 instances).

The networks were usually trained with *Mini-batch Size* of 40. First, a single MCNN was trained on a training set after which it was tested on the testing set. Next, an ensemble of BCCNNs was trained using same data after re-formatting the latter for one-versus-all binary classification. Afterwards the ensemble was tested on the corresponding testing dataset. For the same data, each BCCNN in the ensemble had almost the same training time to train as that of the MCNN.

The results of the experiments are given below.

**Table 1.** *MATLAB Digit Image Dataset* experimentation results

| Network | Training Set Size (NR) | Validation Set Size / Testing Set Size (NS) | Training Epochs | Accuracy (%) |
|---|---|---|---|---|
| MCNN | 1000 | 500 | 8 | 94.00 |
| BCCNN Ensemble | 1000 | 500 | 7 | 93.60 |
| MCNN | 5000 | 1000 | 3 | 99.10 |
| BCCNN Ensemble | 5000 | 1000 | 3 | 99.20 |
| MCNN | 7000 | 1500 | 3 | 99.73 |
| BCCNN Ensemble | 7000 | 1500 | 3 | 99.80 |

**Table 2.** *USPS+ Digit Image Dataset* experimentation results

| Network | Training Set Size (NR) | Validation Set Size /Testing Set Size (NS) | Training Epochs | Accuracy (%) |
|---|---|---|---|---|
| MCNN | 5000 | 1000 | 4 | 98.60 |
| BCCNN Ensemble | 5000 | 1000 | 4 | 98.80 |
| MCNN | 7000 | 1750 | 4 | 99.14 |
| BCCNN Ensemble | 7000 | 1750 | 3 | 99.03 |

**Table 3.** *MNIST Digit Image Dataset* experimentation results #1

| Network | Training Set Size (NR) | Validation Set Size /Testing Set Size (NS) | Training Epochs | Accuracy (%) |
|---|---|---|---|---|
| MCNN | 7000 | 2500 | 5 | 98.44 |
| BCCNN Ensemble | 7000 | 2500 | 4 | 97.44 |
| MCNN | 9000 | 3000 | 3 | 97.90 |
| BCCNN Ensemble | 9000 | 3000 | 4 | 98.03 |

For the BCCNN Ensemble, it was observed that the classification accuracy increased with increase in amount of training data. As shown in Table 3, it was observed that after increasing the training set size from 7000 to 9000, the BCCNN ensemble was able to outperform the MCNN on grounds of classification accuracy. For an MNIST training subset with a size of 9000, a validation set size of 3000 and a testing set size of 3000, the accuracy of the trained BCCNN ensemble was 98.03% which was higher than that of the MCNN viz. 97.90%. One more observed characteristic of BCCNNs was faster training convergence than that of the MCNN.

More experimentation was done on the architecture of BCCNNs in order to improve their recognition accuracy. It was observed that if an eight-neuron fully-connected layer was inserted before the binary layer while retaining all other layers in the network, the recognition accuracy of the modified BCCNN ensemble was higher. For a randomly selected subset of the MNIST Digit Image Dataset, this observation was made as given in Table 4.

**Table 4.** *MNIST Digit Image Dataset* experimentation results #2

| Network | Training Set Size (NR) | Validation Set Size /Testing Set Size (NS) | Training Epochs | Accuracy (%) |
|---|---|---|---|---|
| MCNN | 6000 | 2000 | 3 | 97.75 |
| BCCNN Ensemble | 6000 | 2000 | 6 | 97.65 |
| Modified BCCNN Ensemble | 6000 | 2000 | 5 | 97.80 |

As is observed from Table 4, the ensemble with the modified BCCNNs performed best on the given dataset. It must be noted that each BCCNN in the modified ensemble had two fully connected layers in succession. The first fully connected layer had eight neurons and the second fully connected layer had two neurons. These layers were followed by a softmax layer and a binary sigmoid classification layer respectively in that order.

Further, as suggested by the fast training convergence of the BCCNNs (Figure 2.a), the learning rate was high which might have affected the training of the BCCNNs negatively. In order to investigate the effect of learning rate on training of the BCCNNs, the learning rate for training the ensemble was lowered from 0.01 to 0.005. The results of subsequent testing suggested an improvement in overall classification accuracy. The results of the experiments are shown in Table 5.

**Table 5.** *MNIST Digit Image Dataset* experimentation results #3

| Network | Learning Rate | Training Set Size (NR) | Validation Set Size / Testing Set Size (NS) | Training Epochs | Accuracy (%) |
|---|---|---|---|---|---|

| Network | Learning Rate | Training Set Size (NR) | Validation Set Size /Testing Set Size (NS) | Training Epochs | Accuracy (%) |
|---|---|---|---|---|---|
| MCNN | 0.01 | 3000 | 1250 | 6 | 96.48 |
| BCNN Ensemble | 0.01 | 3000 | 1250 | 5 | 96.80 |
| BCNN Ensemble | 0.005 | 3000 | 1250 | 5 | 97.52 |
| Modified BCNN Ensemble | 0.005 | 3000 | 1250 | 6 | 97.84 |

As is observed from Table 5, the modified BCCNN ensemble having been trained using a learning rate of 0.005 performed best. Each BCCNN in this modified ensemble had two fully connected layers (as detailed previously).

In the context of this work, first, the *MATLAB Digit Image Dataset* as well as the *USPS+ Digit Image Dataset* were used exhaustively. The *MNIST Digit Image Dataset* was also used to some extent. Next, for comprehensive testing of the MNIST Digit Image Dataset, it was also used extensively. From MNIST a randomly selected subset with 40000 training instances, 8000 validation instances and 8000 testing instances respectively, was used. Table 6 gives a comparison of the experimental results for this large subset, after using it on an MCNN and on a modified BCCNN ensemble trained using a low learning rate of 0.005.

**Table 6.** *MNIST Digit Image Dataset* experimentation results #4

| Network | Learning Rate | Training Set Size (NR) | Validation Set Size /Testing Set Size (NS) | Training Epochs | Accuracy (%) |
|---|---|---|---|---|---|
| MCNN | 0.01 | 40000 | 8000 | 2 | 98.4875 |
| Modified BCCNN Ensemble | 0.005 | 40000 | 8000 | 2 | 98.5000 |

As is observed from Table 6, the modified BCCNN ensemble trained using a low learning rate had higher classification accuracy as compared to that of the MCNN. As is observed from the experimentation, using OVA CNN ensembles leads to better classification performance as compared to the traditional multiclass CNN architecture (baseline).

## 5. Conclusion

Using ensembles of One-Versus-All (OVA) deep networks for multiclass classification is a promising area of research. To the best of our knowledge, this is the first work to use an OVA ensemble of CNNs. Each network in the OVA CNN ensemble is trained end-to-end using the traditional OVA-based approach. For testing the trained ensemble, a sample is presented to each network in the ensemble. Prediction score voting of all deep networks in the ensemble is done. After voting, the ensemble member with the largest score is assumed to have classified the sample. For experimentation, the task of digit-image recognition has been used. The experimentation has been done on the *MNIST digit dataset*, the *USPS+ digit dataset*, and on the *MATLAB digit image dataset*. Benchmarking experiments show state-of-art results for the proposed approach. Continuing in this line of research, future work would involve making more modifications to the architecture of BCCNNs in order to increase their efficiency. Also, work would be done on extending the applications of the proposed approach to different areas of deep learning with the help of larger and more powerful deep networks and more complicated tasks like instance segmentation [32], etc.